\documentclass[11pt]{article}

\usepackage[utf8]{inputenc}
\usepackage[T1]{fontenc}
\usepackage{lmodern}
\usepackage{microtype}
\usepackage{geometry}
\usepackage{amsmath,amssymb}
\usepackage{booktabs}
\usepackage{tabularx}
\usepackage{array}
\usepackage{graphicx}
\usepackage{caption}
\usepackage{subcaption}
\usepackage{xcolor}
\usepackage{enumitem}
\usepackage[numbers,sort&compress]{natbib}
\usepackage{hyperref}
\hypersetup{
  colorlinks=true,
  linkcolor=blue,
  citecolor=blue,
  urlcolor=blue,
  pdftitle={A Task-Centric Ontology and Deterministic Domain Rules as a Verifiable Core for AI-Assisted Chemistry Problem Solving},
  pdfauthor={Abduchaborov Ibrokhimsho}
}
\newcommand{\systemname}{\textsc{ChemOntoRule}}
\newcommand{\code}[1]{\texttt{#1}}

\title{A Task-Centric Ontology and Deterministic Domain Rules as a Verifiable Core for AI-Assisted Chemistry Problem Solving}

\author{
Abduchaborov Ibrokhimsho\\
Independent Researcher, Tajikistan\\
\href{mailto:ibrokhimsho.a@gmail.com}{ibrokhimsho.a@gmail.com}\\
ORCID: \href{https://orcid.org/0009-0009-2287-5266}{0009-0009-2287-5266}
}

\date{July 2026}

\begin{document}
\maketitle

\begin{abstract}
Large language models can interpret natural-language chemistry questions, but their internal reasoning is difficult to inspect, constrain, and validate. This paper presents \systemname, a proof-of-concept symbolic core for AI-assisted school-level chemistry problem solving. The central design choice is task-centric ontology engineering: the ontology is constructed around the concepts, properties, relations, and executable procedures required by a defined collection of chemistry problems, rather than as a universal representation of chemistry. The implemented artifact combines a lightweight ontology serialized in JSON and RDF/Turtle with deterministic Python rules for electronic structure, periodic trends, oxidation states, oxide and hydride behavior, and related school-level reasoning patterns. A separate expert-coded fallback handles problem families not yet represented by general rules. The system was examined on 300 human-authored and manually validated chemistry problems. The complete system matched 296 of 300 reference answers (98.67\%). The ontology-driven rule subset covered 269 problems and matched 266 references (98.88\%); 31 problems were handled by task-specific expert-coded fallbacks, with 30 matches. Because the same collection informed ontology construction and evaluation, these results measure implemented coverage and internal consistency, not independent generalization. We analyze the four mismatches, distinguish structural validation from chemical correctness, and define a future architecture in which a language model acts primarily as a translator from user language into a normalized ontological task frame. Token efficiency is presented as a testable hypothesis for future controlled studies, not as a result of the current work.
\end{abstract}


\section{Introduction}

Large language models (LLMs) have demonstrated broad chemical knowledge and strong performance on selected chemistry benchmarks, while also exhibiting inconsistency, overconfidence, and failures on apparently elementary problems \citep{mirza2025chembench,guo2023chemllmbench,wang2024scibench,xie2025qcbench}. These properties make LLMs useful as language interfaces but problematic as the sole source of domain reasoning in settings where intermediate steps, rule applicability, and final answers must be inspectable.

A complementary approach is to represent relevant domain knowledge explicitly and execute deterministic procedures over that representation. Ontologies provide machine-readable classes, properties, individuals, and axioms \citep{gruber1993ontology,w3cowl2}; external rule engines can then use ontology-linked facts to perform calculations or classifications. KnowTD, for example, couples a thermodynamics ontology with a reasoner that selects equations and produces explainable solutions \citep{vollmer2024knowtd}. More broadly, neuro-symbolic research seeks to combine the language flexibility of neural models with the transparency and controllability of symbolic systems \citep{wan2024neurosymbolic,colelough2025review}.

\begin{figure}[t]
  \centering
  \includegraphics[width=\textwidth]{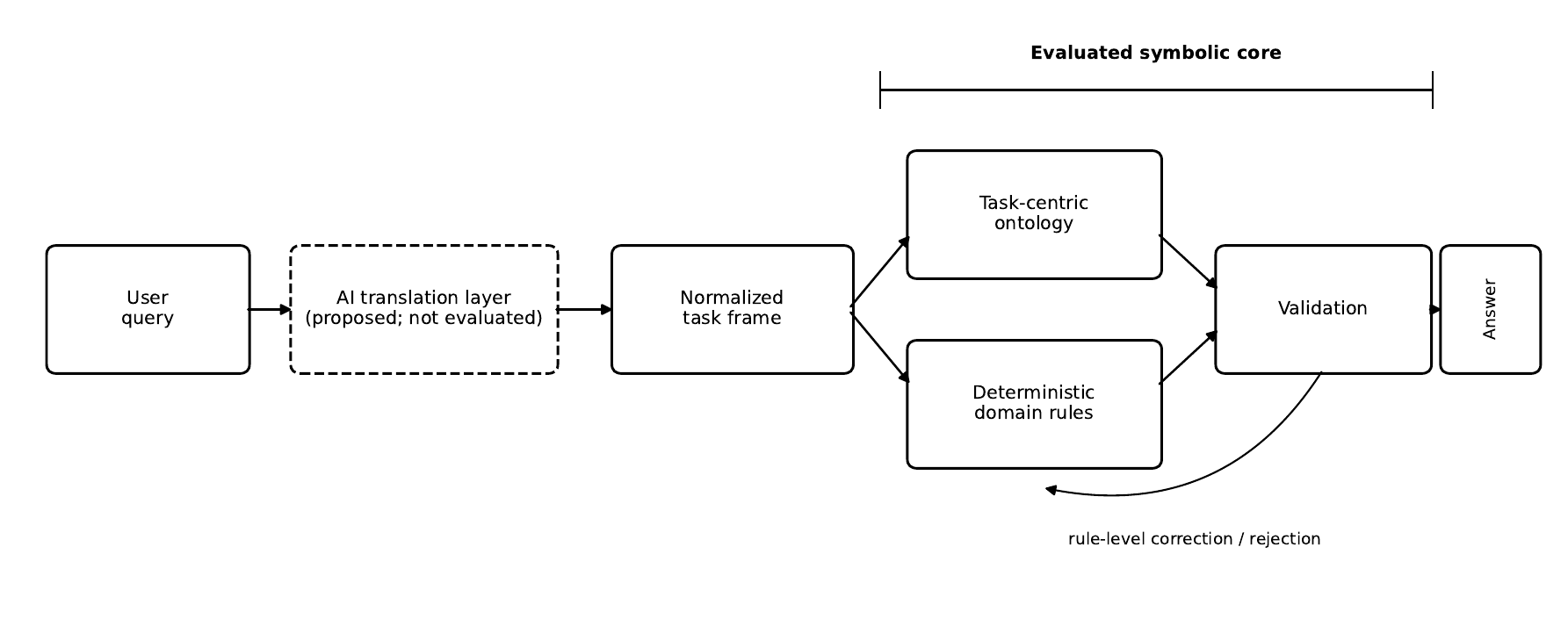}
  \caption{Intended AI-assisted architecture. The dashed AI translation layer is proposed but not evaluated in this paper. The current study evaluates the ontology, deterministic domain rules, reference matching, and structural validation inside the symbolic core.}
  \label{fig:architecture}
\end{figure}

This paper studies a narrower and deliberately practical question: can a task-centric ontology and a deterministic domain-rule engine form a verifiable core for future AI-assisted chemistry problem solving? The intended division of labor is asymmetric. A language model is expected to translate a user's free-form request into a constrained task representation and later verbalize the result; the chemistry reasoning itself is delegated to explicit ontology-linked procedures wherever possible. Figure~\ref{fig:architecture} illustrates this intended architecture and marks the symbolic core evaluated in the current study.

The contribution is not a claim that the system solves chemistry in general. Instead, the paper contributes:

\begin{enumerate}[leftmargin=1.5em]
  \item a task-centric method for deriving ontology content from the competency requirements of a bounded problem collection;
  \item an ontology-backed external rule engine for selected school-level chemistry families;
  \item a transparent separation between general ontology-driven rules and task-specific expert-coded fallbacks;
  \item a proof-of-concept evaluation on 300 human-authored and manually validated problems, including coverage, reference matching, and error analysis;
  \item a carefully scoped proposal for using an LLM as a natural-language translator into the symbolic system, with token efficiency reserved as a future empirical question.
\end{enumerate}

\section{Related Work}

\subsection{LLMs and chemistry problem solving}

ChemLLMBench evaluates LLMs across eight chemistry tasks and shows that model performance varies substantially by task and prompting condition \citep{guo2023chemllmbench}. ChemBench provides a larger framework with more than 2,700 chemistry question-answer pairs and reports both strong average performance and persistent basic errors and overconfidence \citep{mirza2025chembench}. ChemLLM explores domain-specific language-model development for chemistry \citep{zhang2024chemllm}. SciBench evaluates scientific problem solving across mathematics, physics, and chemistry and highlights the difficulty of multi-step scientific reasoning \citep{wang2024scibench}. QCBench focuses on quantitative chemistry and explicitly emphasizes stepwise numerical reasoning across multiple subfields and difficulty levels \citep{xie2025qcbench}.

These works motivate domain-specific evaluation, but the present study differs in objective. It does not primarily benchmark an LLM. Instead, it evaluates a deterministic symbolic core designed to receive a structured task representation from an eventual AI language interface.

\subsection{Ontology-based and neuro-symbolic reasoning}

Ontologies formalize concepts and their relationships and can be exchanged as RDF graphs or represented in OWL \citep{w3cowl2}. Validation languages such as SHACL provide a complementary mechanism for checking RDF graphs against explicit constraints \citep{w3cshacl}. In scientific problem solving, KnowTD demonstrates an actionable knowledge representation in which an ontology and reasoner cooperate to select thermodynamic equations and solve introductory problems \citep{vollmer2024knowtd}. Comparative work on LLMs and thermodynamic problem solving further illustrates both the flexibility and variability of generative approaches \citep{loubet2025thermo}.

Recent work on ontology-guided language-model inference argues that explicit domain knowledge can improve reliability when retrieval and grounding are accurate, while irrelevant context may degrade performance \citep{labre2026ontology}. Neuro-symbolic surveys similarly emphasize interpretability, robustness, and explicit reasoning as motivations for hybrid architectures \citep{wan2024neurosymbolic,colelough2025review}.

\subsection{Task-centric ontology construction}

Ontology engineering commonly uses competency questions to state what the ontology must be able to answer. Recent work has investigated LLM support for generating such questions, while retaining the central role of domain requirements and expert assessment \citep{pan2024competency}. The present system follows a pragmatic competency-driven process: problem families are inspected; required entities, properties, and rules are identified; and only the knowledge necessary for the target task space is implemented. This produces a narrower artifact than a comprehensive chemistry ontology, but one that is directly executable for a defined set of problem patterns.

\section{Research Scope and Questions}

The study addresses four questions:

\begin{description}[leftmargin=2.8em,style=nextline]
  \item[RQ1: Coverage.] What proportion of the 300-problem collection can be routed to general ontology-driven rules rather than task-specific fallbacks?
  \item[RQ2: Reference matching.] How often do the complete system, the ontology-rule subset, and the fallback subset match the human-validated reference answers?
  \item[RQ3: Failure modes.] Which modeling or implementation assumptions explain the observed mismatches?
  \item[RQ4: AI integration.] What interface should a future LLM-based translator expose so that natural-language requests can be resolved by the deterministic symbolic core?
\end{description}

The paper does \emph{not} claim out-of-distribution generalization, superiority over a controlled LLM baseline, reduced inference cost, or measured token savings.

\section{Problem Collection and Validation}

The evaluation collection contains 300 human-authored school-level chemistry problems accompanied by reference answers and solutions. The collection was manually reviewed and validated for use in this study. Full task statements are not redistributed because they may be subject to third-party licensing restrictions. The released manuscript package therefore reports aggregate statistics, system design, and mismatch descriptions without reproducing the complete problem text.

The collection served two purposes: it supplied requirements for the task-centric ontology and it provided a proof-of-concept evaluation set. This dual use is central to the interpretation of the results. The reported numbers characterize implemented coverage and internal consistency within the designed task space; they do not estimate performance on an independent unseen distribution.

Table~\ref{tab:answer-types} shows that the collection is dominated by selection-sequence questions. This imbalance limits conclusions about open-ended derivations and numeric problem solving.

\begin{table}[t]
\centering
\caption{Answer-type distribution in the 300-problem collection.}
\label{tab:answer-types}
\begin{tabular}{lr}
\toprule
Answer type & Number of problems \\
\midrule
Selection sequence & 291 \\
Equation set & 5 \\
Scalar answer & 3 \\
Structured calculation object & 1 \\
\midrule
Total & 300 \\
\bottomrule
\end{tabular}
\end{table}

\begin{figure}[t]
  \centering
  \includegraphics[width=0.82\textwidth]{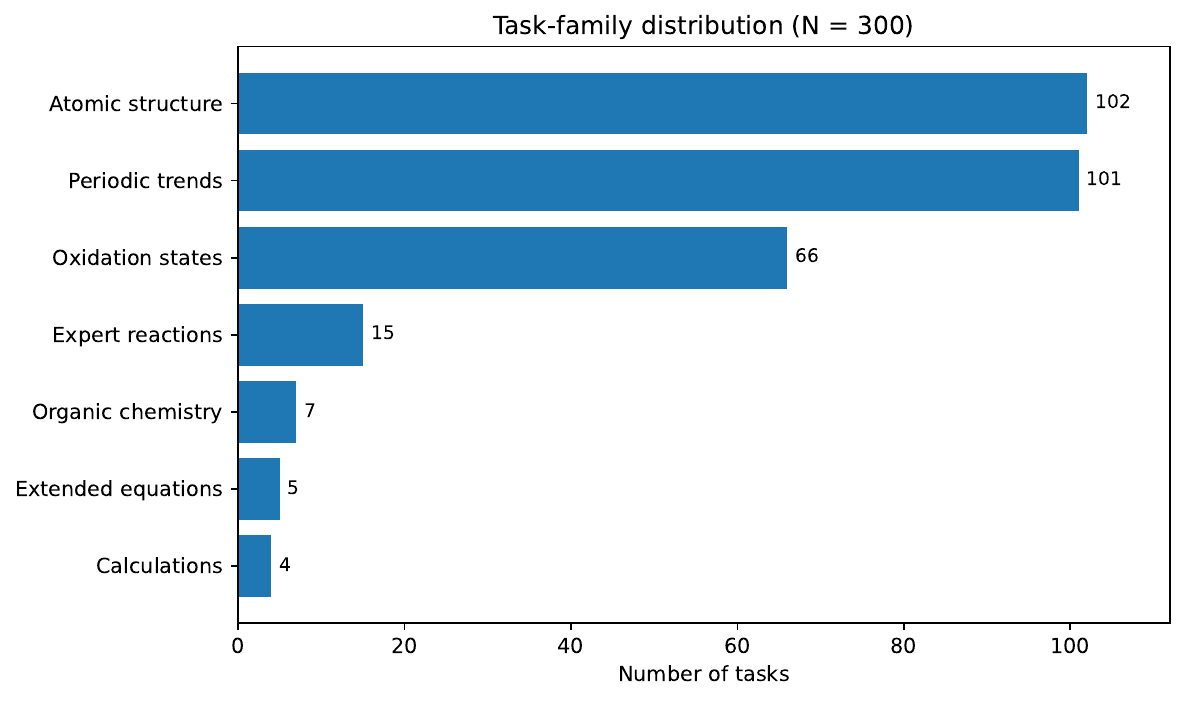}
  \caption{Distribution of the 300 problems across implementation-level task families.}
  \label{fig:distribution}
\end{figure}

The average problem length was 217.7 characters and the solution records contained an average of 2.24 linked concepts. The seven operational task families are shown in Figure~\ref{fig:distribution}. These are routing categories used by the implementation and are mutually exclusive in the reported metrics.

\section{Task-Centric Ontology}

\subsection{Design principle}

A conventional domain ontology may aim to represent chemistry broadly. The present ontology instead begins with a bounded task space. For each recurring problem family, the development process identifies:

\begin{enumerate}[leftmargin=1.5em]
  \item the entities and values explicitly present in the question;
  \item the target relation, classification, or ordering;
  \item the chemical properties required to reach the target;
  \item the conditions under which a rule is applicable;
  \item the output format and available structural checks.
\end{enumerate}

This process can be interpreted as a set of competency questions. Examples include: Which elements have a specified number of electrons on the outer level? Which oxidation states are represented by a given oxide formula? Which selected elements should be ordered by a particular periodic trend? Which ion-electron transformation yields a stable shell configuration?

\subsection{Formal view}

Let the ontology be

\begin{equation}
\mathcal{O} = (\mathcal{C}, \mathcal{P}, \mathcal{I}, \mathcal{A}),
\end{equation}

where $\mathcal{C}$ is a set of classes, $\mathcal{P}$ a set of properties and relations, $\mathcal{I}$ a set of individuals, and $\mathcal{A}$ a set of asserted or derived attribute values. An executable domain rule is represented abstractly as

\begin{equation}
r_j: \phi_j(z,\mathcal{O}) \rightarrow (y_j,\tau_j),
\end{equation}

where $z$ is a normalized task frame, $\phi_j$ is an applicability predicate, $y_j$ is a result, and $\tau_j$ is a trace or explanation fragment.

For a user query $q$, the intended AI translator $T_\theta$ will eventually produce

\begin{equation}
z = T_\theta(q) = (f, E, K, o),
\end{equation}

where $f$ denotes the task family, $E$ the recognized entities, $K$ the constraints or requested property, and $o$ the output schema. The deterministic core then computes

\begin{equation}
S(z,\mathcal{O},\mathcal{R}) \rightarrow (a,\tau,v),
\end{equation}

with answer $a$, trace $\tau$, and validation status $v$. The current study evaluates $S$ using constrained deterministic parsing of the supplied task formats; it does not evaluate $T_\theta$.

\subsection{Implemented ontology content}

The working ontology contains four declared conceptual classes: \code{Element}, \code{PeriodicTrend}, \code{OxidationStateRule}, and \code{ManualReactionPattern}. Eight task-relevant relation groups connect elements to periods, groups, blocks, outer-level structure, oxidation states, oxide formation, volatile-hydride formation, and trend scores. The instance layer includes 52 chemical elements with stored and derived attributes. The RDF/Turtle serialization contains approximately 1,029 triples, including 17 datatype properties.

The ontology is intentionally lightweight. Most executable logic remains in an external Python rule engine rather than in OWL axioms, SWRL rules, or SHACL functions. Accordingly, the implementation is best described as an \emph{ontology-backed external rule engine}, not a pure OWL reasoner.

\begin{figure}[t]
  \centering
  \includegraphics[width=0.92\textwidth]{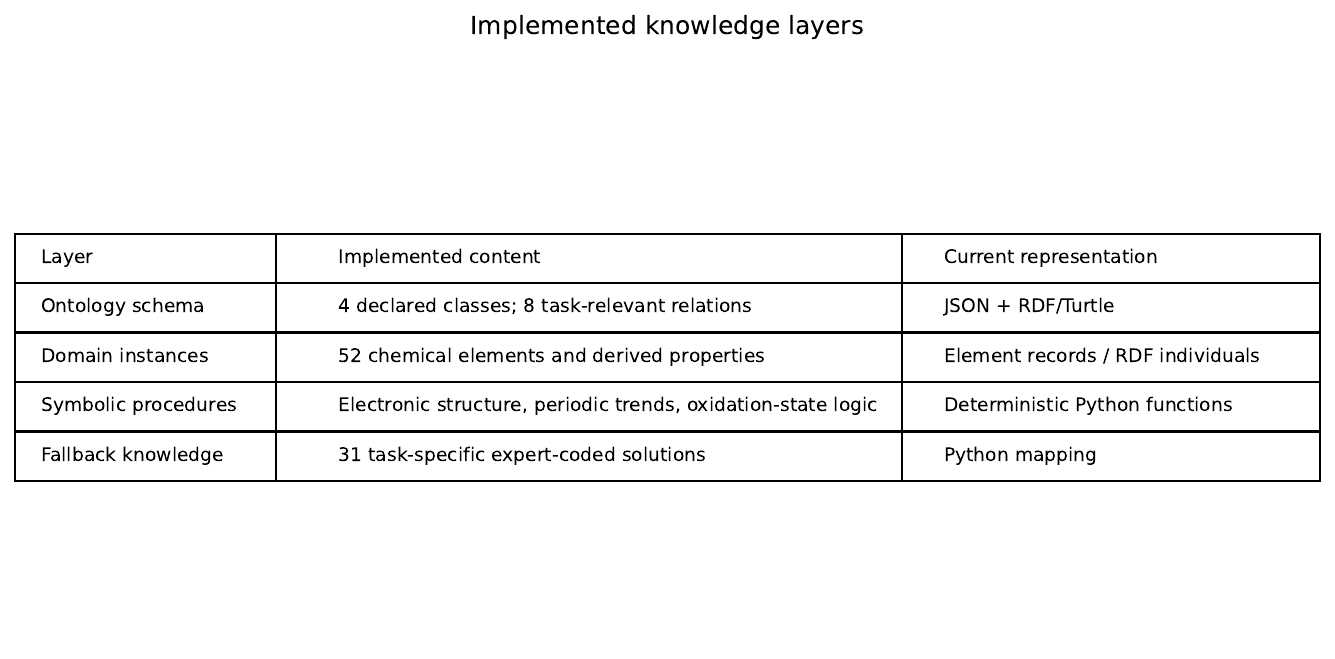}
  \caption{Knowledge layers in the current implementation. General symbolic procedures and task-specific expert fallbacks are separated explicitly.}
  \label{fig:layers}
\end{figure}

\section{Deterministic Domain-Rule Engine}

The rule engine is implemented in Python and derives or evaluates school-level chemical properties using explicit procedures. The principal groups are:

\begin{itemize}[leftmargin=1.5em]
  \item electron-configuration construction and subshell counting;
  \item outer-level and valence-electron calculations;
  \item unpaired-electron calculations;
  \item stable-ion electron counts;
  \item highest, lowest, and ranges of oxidation states;
  \item feasibility checks for common oxide formula patterns;
  \item heuristic scores for metallic, nonmetallic, acidic, basic, and hydride trends;
  \item routing and ordering rules for recognized task templates.
\end{itemize}

Each solved record includes a final answer, linked concepts, a short solution trace, and a method label. The method label is either \code{ontology\_rule} or \code{manual\_expert}. The first denotes a reusable deterministic procedure linked to the ontology. The second denotes a task-specific expert-coded solution used when the general rule base does not cover the problem family.

This distinction prevents fallback answers from being misrepresented as ontology inference. It also exposes a concrete expansion path: a fallback problem family becomes general only after its entities, applicability conditions, transformations, and validation rules are represented independently of task identifiers.

\section{Processing and Validation}

The instrumentation represents each solution as six logical stages:

\begin{enumerate}[leftmargin=1.5em]
  \item task intake;
  \item condition parsing;
  \item solver routing;
  \item ontology-context lookup;
  \item answer production;
  \item result validation.
\end{enumerate}

Across 300 tasks, this yields 1,800 stage records and 1,800 event logs. The stage durations in the current dashboard are generated for visualization and are not wall-clock measurements. They must not be interpreted as latency results.

The built-in validation checks output structure: presence of an answer, non-empty selections, plausible selection length, and option-index range. It does not independently establish chemical correctness. We therefore report two distinct notions:

\begin{description}[leftmargin=2.8em,style=nextline]
  \item[Structural validity] Whether the output conforms to expected format constraints.
  \item[Reference match] Whether the normalized output equals the human-validated reference answer.
\end{description}

All 300 outputs passed the implemented structural checks. Chemical performance is reported using reference matching.

\section{Experimental Protocol}

The solver was run over all 300 problems. For each task, the system stored the route, family, answer type, concept count, final answer, and solution trace. Reference matching used normalized exact comparison. The principal reported measures are:

\begin{equation}
\mathrm{Coverage}_{\mathrm{rules}} = \frac{N_{\mathrm{ontology\_rule}}}{N_{\mathrm{all}}},
\end{equation}

\begin{equation}
\mathrm{MatchRate} = \frac{N_{\mathrm{matched}}}{N_{\mathrm{evaluated}}}.
\end{equation}

Wilson 95\% confidence intervals are included as descriptive uncertainty intervals for the observed proportions. Because the collection was used during development, these intervals do not correct for design-set dependence and should not be interpreted as population-level generalization bounds.

A separate file of LLM-generated answers was available for descriptive comparison. It matched all 300 reference answers. However, the exact model version, full prompts, decoding parameters, token counts, repeated-run protocol, and possible post-processing history were not preserved. Consequently, the LLM file is not treated as a controlled baseline and no superiority claim or statistical comparison is made.

\section{Results}

\subsection{Coverage and reference matching}

The ontology-driven rule engine covered 269 of 300 problems, corresponding to 89.67\% rule coverage. The remaining 31 problems were processed by task-specific expert-coded fallbacks.

\begin{figure}[t]
  \centering
  \includegraphics[width=0.78\textwidth]{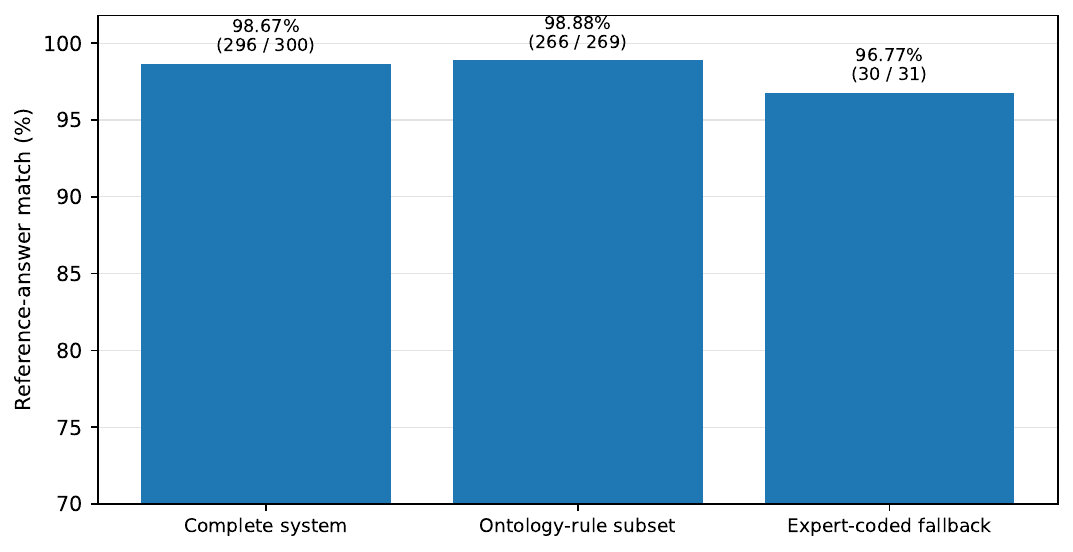}
  \caption{Reference-answer matching for the complete system and the two routing subsets. These are internal proof-of-concept results, not estimates of unseen-task generalization.}
  \label{fig:results}
\end{figure}

The complete system matched 296 of 300 reference answers (98.67\%). The ontology-rule subset matched 266 of 269 references (98.88\%), while the fallback subset matched 30 of 31 (96.77\%). Table~\ref{tab:main-results} and Figure~\ref{fig:results} summarize these results.

\begin{table}[t]
\centering
\caption{Proof-of-concept reference matching. Confidence intervals are Wilson 95\% intervals for the observed proportions.}
\label{tab:main-results}
\begin{tabular}{lrrrr}
\toprule
Component & Evaluated & Matched & Match rate & 95\% CI \\
\midrule
Complete system & 300 & 296 & 98.67\% & [96.62, 99.48] \\
Ontology-rule subset & 269 & 266 & 98.88\% & [96.77, 99.62] \\
Expert-coded fallback & 31 & 30 & 96.77\% & [83.81, 99.43] \\
\bottomrule
\end{tabular}
\end{table}

\subsection{Results by task family}

Table~\ref{tab:family-results} shows the family-level outcomes. All oxidation-state, expert-reaction, organic, and extended-equation tasks matched the references. Three of the four mismatches occurred in general ontology-rule families, and one occurred in an expert-coded calculation.

\begin{table}[t]
\centering
\caption{Reference matching by implementation-level task family.}
\label{tab:family-results}
\begin{tabular}{lrrr}
\toprule
Task family & Tasks & Matched & Match rate \\
\midrule
Atomic structure & 102 & 100 & 98.04\% \\
Periodic trends & 101 & 100 & 99.01\% \\
Oxidation states & 66 & 66 & 100.00\% \\
Expert reactions & 15 & 15 & 100.00\% \\
Organic chemistry & 7 & 7 & 100.00\% \\
Extended equations & 5 & 5 & 100.00\% \\
Calculations & 4 & 3 & 75.00\% \\
\midrule
Total & 300 & 296 & 98.67\% \\
\bottomrule
\end{tabular}
\end{table}

\section{Mismatch Analysis}

Four outputs did not match the references. Table~\ref{tab:mismatches} presents a compact diagnosis. The purpose of this analysis is not only to repair individual tasks but to identify which abstractions should be added or corrected in the reusable rule layer.

\begin{table*}[t]
\centering
\caption{Analysis of the four reference mismatches.}
\label{tab:mismatches}
\begin{tabularx}{\textwidth}{p{1.25cm}p{2.1cm}p{2.1cm}X}
\toprule
Task ID & Reference & System output & Diagnosis \\
\midrule
314 & 19\% BaCl$_2$; 3.85\% Ba(OH)$_2$ & 19.2\% BaCl$_2$; 3.6\% Ba(OH)$_2$ & The expert-coded calculation contains a numerical derivation inconsistent with the reference. The case should be replaced by a general stoichiometric and mass-fraction procedure with explicit equation checks and tolerance rules. \\
3129 & 34 & 35 & The rule selected elements with three outer-level electrons but did not represent the full condition ``lose three electrons to obtain an eight-electron outer shell.'' The distinction between outer-electron count and the resulting stable shell must be explicit. \\
3149 & 13 & 23 & The implementation compared aggregate $p$- and $s$-subshell electron counts rather than counts on the outer level. The ontology query and rule scope must bind both quantities to the same principal energy level. \\
3485 & 421 & 412 & A simple trend heuristic was insufficient for ordering reducing properties of selected transition metals. The rule requires a more appropriate school-level ordering source or an electrochemical/domain-specific relation rather than a generic metallicity score. \\
\bottomrule
\end{tabularx}
\end{table*}

These errors illustrate three distinct classes of weakness:

\begin{enumerate}[leftmargin=1.5em]
  \item \textbf{missing semantic scope}, as in ``outer level'' versus all occupied subshells;
  \item \textbf{incomplete applicability conditions}, as in stable-shell formation after electron loss;
  \item \textbf{over-simplified domain heuristics}, especially for transition-metal behavior;
  \item \textbf{task-specific arithmetic}, which should be replaced by reusable equations and validators.
\end{enumerate}

The first two errors are especially instructive for ontology engineering. A property may be numerically available yet semantically wrong for the question if its scope is not modeled. Explicit relations such as \code{belongsToPrincipalLevel} and constraints requiring the same outer level would reduce this ambiguity.

\section{AI Translation Layer}

The long-term architecture treats an LLM as a translator and interaction layer, not as the final chemistry reasoner. The proposed translator receives a user's natural-language question and emits a normalized task frame such as:

\begin{verbatim}
{
  "task_family": "atomic_structure",
  "entities": ["O", "Si", "Cl", "H", "Li"],
  "requested_relation": "outer_p_electrons > outer_s_electrons",
  "constraints": {"state": "ground"},
  "output_schema": "two_option_indices"
}
\end{verbatim}

The symbolic core then selects an applicable rule, executes it, records the trace, and validates the output. A final language component may verbalize the trace for the user. This division offers three conceptual advantages:

\begin{itemize}[leftmargin=1.5em]
  \item the natural-language interface can accept varied wording without embedding all chemistry logic in prompts;
  \item domain reasoning remains inspectable and amendable by experts;
  \item rule failures can be localized to parsing, ontology mapping, applicability, execution, or validation.
\end{itemize}

The present experiment does not implement or evaluate this LLM translator. The architecture is included because it defines the intended role of AI and determines the interface requirements of the symbolic core.

\section{Discussion}

\subsection{What the result demonstrates}

The experiment shows that a relatively small, task-focused ontology combined with deterministic procedures can cover a large fraction of a bounded problem collection. The 269-rule coverage figure is meaningful as an engineering result: most tasks were handled through reusable functions rather than direct answer lookup. The four mismatches also show that explicit systems do not eliminate errors; instead, they make errors local and inspectable.

The distinction between ontology data and executable rules is important. The current RDF/Turtle graph stores element facts and properties, while Python performs most inference. This separation is practical for a prototype, but it means the term ``ontology reasoning'' should not be interpreted as OWL-only logical inference. A more formal future version could move validation and selected derivations into SHACL, SHACL rules, SWRL, or another declarative rule language.

\subsection{Why task-centric design is useful}

Task-centric design limits scope intentionally. It avoids the cost of constructing a complete chemistry ontology before any task can be solved. It also provides a direct expansion procedure: collect unsupported tasks, identify recurring competency requirements, add ontology distinctions and general rules, and then remove task-specific fallbacks when equivalent coverage is achieved.

The same design creates a methodological risk. If ontology content is derived from the evaluation collection, high internal match rates may reflect careful engineering around known patterns rather than transfer to unseen tasks. For this reason, the system should be described as a proof of concept and an implemented-coverage study.

\subsection{AI and possible token efficiency}

The architecture suggests, but does not demonstrate, a token-efficiency hypothesis. If an LLM only maps user language into a compact task frame and verbalizes a deterministic trace, it may require fewer prompt tokens and fewer generated reasoning tokens than an LLM asked to reproduce all domain reasoning in natural language. It may also reduce repeated correction calls after chemically invalid generations.

A future controlled experiment should compare at least four configurations: LLM-only, LLM with textual retrieval, LLM with ontology context, and LLM translator plus symbolic execution. It should record input, output, cached, and reasoning tokens; all retries; tool calls; latency; cost; and accuracy. The key measure should be tokens or cost per correct solution, not merely average response length. No token-saving claim is made in the present paper.

\section{Threats to Validity and Limitations}

\paragraph{Development--evaluation overlap.}
The problem collection was used to derive requirements and evaluate the artifact. The reported rates therefore measure internal coverage and consistency, not generalization.

\paragraph{Task distribution.}
Selection-sequence questions constitute 291 of 300 cases. The evidence for open-ended equations, calculations, and explanatory reasoning is limited.

\paragraph{Expert-coded fallbacks.}
Thirty-one tasks use task-specific expert-coded solutions. Their performance must be reported separately and cannot be treated as general ontology-driven inference.

\paragraph{Lightweight ontology.}
Most rules are procedural Python logic outside the RDF/OWL graph. The system does not currently exploit the full expressive or validation capabilities of OWL and SHACL.

\paragraph{Reference-answer evaluation.}
Normalized exact matching does not verify every intermediate chemical step. Conversely, a reference mismatch may sometimes reflect rounding conventions or ambiguity. Expert adjudication should accompany future independent tests.

\paragraph{LLM comparison.}
The available LLM answer file lacks sufficient run metadata for a controlled benchmark. It is excluded from the central claims.

\paragraph{No performance or token measurements.}
Dashboard durations are visualization values rather than real execution times. Token usage, API cost, and wall-clock latency were not measured.

\paragraph{Data redistribution.}
Full problem texts are not included because they may be subject to third-party licensing restrictions, limiting direct replication from the manuscript package alone.

\section{Future Work}

The next research steps are:

\begin{enumerate}[leftmargin=1.5em]
  \item create a locked independent test set containing new wording, new entities, and new compositions of known rules;
  \item implement the LLM translator from free-form user requests to normalized ontology frames and evaluate entity linking, task routing, and constraint extraction separately;
  \item replace task-specific fallbacks with reusable reaction, calculation, and organic-chemistry modules;
  \item add declarative constraints using SHACL and investigate which procedural rules can be represented in a formal rule language;
  \item collect real latency, memory, token, and cost logs across controlled AI baselines;
  \item evaluate robustness to paraphrases, irrelevant details, missing information, and intentionally inconsistent questions;
  \item publish an author-created and openly licensed benchmark suitable for independent replication.
\end{enumerate}

\section{Conclusion}

This paper presented a task-centric chemistry ontology and deterministic domain-rule engine as a verifiable core for future AI-assisted problem solving. The prototype routes 269 of 300 human-authored and manually validated school-level problems to reusable ontology-linked rules and uses expert-coded fallbacks for 31. The complete system matches 296 reference answers, while the ontology-rule subset matches 266 of 269. These results demonstrate implemented coverage and internal consistency within a bounded task space; they do not establish independent generalization.

The principal architectural claim is intentionally modest: an AI system need not perform all chemistry reasoning inside a language model. A language model can instead translate a user's request into an explicit task frame, after which a symbolic core applies inspectable rules and validators. The current work establishes and audits that core. Independent evaluation, formalized constraints, a measured LLM interface, and controlled token-efficiency experiments remain future work.

\section*{Reproducibility and Availability}

\paragraph{Code availability.}
The source code is planned for public release under the Apache License 2.0. Repository details will be announced by the author.

\paragraph{Data availability.}
Aggregate metrics and descriptions of the four mismatches are included in the manuscript package. Full problem statements are not redistributed because they may be subject to third-party licensing restrictions.

\paragraph{Funding.}
This research received no external funding.

\paragraph{Competing interests.}
The author declares no competing interests.

\paragraph{Author contributions.}
Abduchaborov Ibrokhimsho: conceptualization, ontology design, methodology, software, validation, analysis, visualization, and writing.

\appendix

\section{Implementation Summary}

The implementation artifacts used for this study include a JSON ontology, an RDF/Turtle serialization, a deterministic Python solver, solved-task records, aggregate process metrics, stage traces, event logs, and a static dashboard. The ontology metadata describes the scope as atomic structure, periodic properties, oxidation states, oxides and hydrides, plus manual rules for mixed assignments.

\begin{table}[h]
\centering
\caption{Core implementation statistics.}
\begin{tabular}{lr}
\toprule
Statistic & Value \\
\midrule
Problems & 300 \\
Ontology-rule routes & 269 \\
Expert-coded fallback routes & 31 \\
Declared ontology classes & 4 \\
Task-relevant relation groups & 8 \\
Element individuals & 52 \\
Approximate RDF triples & 1,029 \\
Datatype properties in Turtle serialization & 17 \\
Logical processing stages per task & 6 \\
Recorded stage events & 1,800 \\
\bottomrule
\end{tabular}
\end{table}

\section{Descriptive LLM Comparison}

A stored file of LLM-produced solutions matched the normalized reference answers for all 300 tasks, while the symbolic system matched 296. This observation is not used as evidence that the LLM is superior because the experiment lacks the metadata required for a controlled comparison: model version, provider, prompts, decoding settings, number of attempts, token counts, and complete post-processing history. The file is therefore treated only as contextual material.

\section{Recommended Independent Evaluation Design}

A future evaluation should freeze the ontology and rules before revealing a new test set. The test set should contain at least three strata:

\begin{enumerate}[leftmargin=1.5em]
  \item \textbf{paraphrase transfer}: known rule structures with substantially different language;
  \item \textbf{entity transfer}: known rules applied to different elements, substances, or numerical values;
  \item \textbf{compositional transfer}: new multi-step combinations of individually implemented rules.
\end{enumerate}

All AI configurations should receive identical task texts and answer-format constraints. Evaluation should retain every run rather than selecting the best attempt. Human adjudicators should inspect disagreements and classify errors into parsing, ontology mapping, rule applicability, execution, validation, and explanation categories.


\begin{thebibliography}{99}

\bibitem[Mirza et~al.(2025)]{mirza2025chembench}
A. Mirza et al.
\newblock Are large language models superhuman chemists?
\newblock \emph{Nature Chemistry}, 2025.
\newblock doi:10.1038/s41557-025-01815-x; arXiv:2404.01475.

\bibitem[Wang et~al.(2024)]{wang2024scibench}
X. Wang, Z. Hu, P. Lu, Y. Zhu, J. Zhang, S. Subramaniam, A. R. Loomba,
S. Zhang, Y. Sun, and W. Wang.
\newblock SciBench: Evaluating college-level scientific problem-solving abilities of large language models.
\newblock In \emph{Proceedings of the 41st International Conference on Machine Learning}, 2024.
\newblock arXiv:2307.10635.

\bibitem[Guo et~al.(2023)]{guo2023chemllmbench}
T. Guo, K. Guo, B. Nan, Z. Liang, Z. Guo, N. V. Chawla, O. Wiest, and X. Zhang.
\newblock What can large language models do in chemistry? A comprehensive benchmark on eight tasks.
\newblock In \emph{Advances in Neural Information Processing Systems}, volume 36, 2023.
\newblock arXiv:2305.18365.

\bibitem[Zhang et~al.(2024)]{zhang2024chemllm}
D. Zhang et al.
\newblock ChemLLM: A chemical large language model.
\newblock arXiv:2402.06852, 2024.

\bibitem[Xie et~al.(2025)]{xie2025qcbench}
J. Xie, W. Wang, B. Gao, Z. Yang, H. Wan, S. Zhang, T. Fu, and Y. Li.
\newblock QCBench: Evaluating large language models on domain-specific quantitative chemistry.
\newblock \emph{Journal of Chemical Information and Modeling}, 2025.
\newblock doi:10.1021/acs.jcim.5c02033; arXiv:2508.01670.

\bibitem[Vollmer et~al.(2024)]{vollmer2024knowtd}
L. Vollmer, S. Fellenz, F. Jirasek, H. Leitte, and H. Hasse.
\newblock KnowTD---An actionable knowledge representation system for thermodynamics.
\newblock arXiv:2407.17169, 2024.

\bibitem[Loubet et~al.(2025)]{loubet2025thermo}
R. Loubet et al.
\newblock Using large language models for solving thermodynamic problems.
\newblock arXiv:2502.05195, 2025.

\bibitem[Wan et~al.(2024)]{wan2024neurosymbolic}
Z. Wan, C.-K. Liu, H. Yang, C. Li, H. You, Y. Fu, C. Wan, T. Krishna, Y. Lin, and A. Raychowdhury.
\newblock Towards cognitive AI systems: A survey and prospective on neuro-symbolic AI.
\newblock arXiv:2401.01040, 2024.

\bibitem[Colelough and Regli(2025)]{colelough2025review}
B. C. Colelough and W. Regli.
\newblock Neuro-symbolic AI in 2024: A systematic review.
\newblock arXiv:2501.05435, 2025.

\bibitem[Pan et~al.(2024)]{pan2024competency}
X. Pan, J. van Ossenbruggen, V. de Boer, and Z. Huang.
\newblock A RAG approach for generating competency questions in ontology engineering.
\newblock arXiv:2409.08820, 2024.

\bibitem[Labre(2026)]{labre2026ontology}
M. Labre.
\newblock Ontology-guided neuro-symbolic inference: Grounding language models with mathematical domain knowledge.
\newblock arXiv:2602.17826, 2026.

\bibitem[Gruber(1993)]{gruber1993ontology}
T. R. Gruber.
\newblock A translation approach to portable ontology specifications.
\newblock \emph{Knowledge Acquisition}, 5(2):199--220, 1993.

\bibitem[W3C OWL Working Group(2012)]{w3cowl2}
W3C OWL Working Group.
\newblock OWL 2 Web Ontology Language Document Overview (Second Edition).
\newblock W3C Recommendation, 2012.
\newblock \url{https://www.w3.org/TR/owl2-overview/}.

\bibitem[W3C Data Shapes Working Group(2017)]{w3cshacl}
W3C Data Shapes Working Group.
\newblock Shapes Constraint Language (SHACL).
\newblock W3C Recommendation, 2017.
\newblock \url{https://www.w3.org/TR/shacl/}.

\end{thebibliography}
\end{document}